\documentclass[runningheads,english]{llncs}
\usepackage{makeidx}
\usepackage{graphicx}
\usepackage{amsmath,amssymb}
\usepackage{bm}
\usepackage{babel}
\usepackage{color}
\usepackage{url}
\usepackage[unicode=true, pdfusetitle,
 bookmarks=true,bookmarksnumbered=false,bookmarksopen=false,
 breaklinks=false,pdfborder={0 0 1},backref=page,colorlinks=true]{hyperref}

\begin{document}
\pagestyle{headings}
\mainmatter

\title{Gaussian Mixture Modeling with Gaussian Process Latent Variable Models}
\titlerunning{Gaussian Mixture Modeling with GPLVMs}

\authorrunning{Hannes Nickisch and Carl E. Rasmussen}
\author{Hannes Nickisch$^1$ and Carl Edward Rasmussen$^{2,1}$ \\
hn@tue.mpg.de and cer54@cam.ac.uk \hspace{11mm}
}
\institute{$^1$ MPI for Biological Cybernetics, T\"ubingen, Germany \\
$^2$ Department of Engineering, University of Cambridge, UK}

\maketitle

\begin{abstract}
Density modeling is notoriously difficult for high dimensional data. One approach to the problem is to search for a lower dimensional manifold which captures the main characteristics of the data. Recently, the Gaussian Process Latent Variable Model (GPLVM) has successfully been used to find low dimensional manifolds in a variety of complex data. The GPLVM consists of a set of points in a low dimensional latent space, and a stochastic map to the observed space. We show how it can be interpreted as a density model in the observed space. However, the GPLVM is not trained as a density model and therefore yields bad density estimates. We propose a new training strategy and obtain improved generalisation performance and better density estimates in comparative evaluations on several benchmark data sets.
\end{abstract}

Modeling of densities, aka unsupervised learning, is one of the central
problems in machine learning. Despite its long history \cite{izenman91nonparmaDE},
density modeling remains a challenging task especially in high dimensional
spaces. For example, the generative approach to classification requires
density models for each class, and training such models well is generally
considered more difficult than the alternative discriminative approach.
Classical approaches to density modeling include both parametric and
non parametric methods. In general, simple parametric approaches have
limited utility, as the assumptions might be too restrictive. Mixture
models, typically trained using the EM algorithm, are more flexible,
but e.g. Gaussian mixture models are hard to fit in high dimensions,
as each component is either diagonal or has in the order of $D^{2}$
parameters, although the mixtures of Factor Analyzers algorithm \cite{GhahramaniBeal00}
may be able to strike a good balance. Methods based on kernel density
estimation \cite{rosenblatt56DE,parzen62DE} are another approach,
where bandwidths may be set using cross validation \cite{rudemo82CVkde}.

The methods mentioned so far have two main shortcomings: 1) they typically
do not perform well in high dimensions, and 2) they do not provide
an intuitive or generative understanding of the data. Generally, we
can only succeed if the data has some regular structure, the model
can discover and exploit. One attempt to do this is to assume that
the data points in the high dimensional space lie on -- or close to
-- some smooth underlying lower dimensional manifold. Models based
on this idea can be divided into models based on \emph{implicit}
or \emph{explicit} representations of the manifold. An implicit representation
is used by \cite{vincent03manifold} in a non-parametric Gaussian
mixture with adaptive covariance to every data point.
Explicit representations
are used in the Generative Topographic Map \cite{bishop98GTM} and
by \cite{Roweis02globLinMod}. Within the explicit camp, models contain
two separate parts, a lower dimensional latent space equipped with
a density, and a function which maps points from the low dimensional
latent space to the high dimensional space where the observations
lie. Advantages of this type of model include the ability to understand
the structure of the data in a more intuitive way using the latent
representation, as well as the technical advantage that the density
in the observed space is automatically properly normalised by construction.

The Gaussian Process Latent Variable Model (GPLVM) \cite{Lawrence05GPLVM}
uses a Gaussian process (GP) \cite{Rasmussen06GPML} to define a (stochastic)
map between a lower dimensional latent space and the observation space.
However, the GPLVM does not include a density in the latent space.
In this paper, we explore extensions to the GPLVM based on
densities in the latent space. One might assume that this can trivially
be done, by thinking of the latent points learnt by the GPLVM as representing
a mixture of delta functions in the latent space. Since the GP based
map is stochastic, it induces a proper mixture in the
observed space. However, this formulation is unsatisfactory, because
the resulting model is not trained as a density model.
Consequently, our experiments show poor density estimation performance.

Mixtures of Gaussians form the basis
of the vast majority of density estimation algorithms. Whereas kernel
smoothing techniques can be seen as introducing a mixture component
for each data point, infinite mixture models \cite{Rasmussen00}
explore the limit as the number of components increases and mixtures
of factor analysers impose constraints on the covariance of individual
components. The algorithm presented in this paper can be understood
as a method for stitching together Gaussian mixture components in
a way reminiscent of \cite{Roweis02globLinMod} using the GPLVM map
from the lower dimensional manifold to induce factor analysis like
constraints in the observation space. In a nutshell, we propose a
density model in high dimensions by transforming a set of low-dimensional
Gaussians with a GP.

We begin by a short introduction to the GPLVM and show how it can
be used to define density models. In section \ref{sec:learning},
we introduce a principled learning algorithm, and experimentally evaluate
our approach in section \ref{sec:experiments}.

\section{The GPLVM as a Density Model\label{sec:DGPLVM}}

A GP $f$ is a probabilistic
map parametrised by a covariance $k(\mathbf{x},\mathbf{x}')$ and
a mean $m(\mathbf{x})$. We use $m(\mathbf{x})\equiv0$
and automatic relevance determination (ARD)
$k(\mathbf{x}^{i},\mathbf{x}^{j})=\sigma_{f}^{2}\exp\big(\!-\frac{1}{2}(\mathbf{x}^{i}-\mathbf{x}^{j})^{\top}\mathbf{W}^{-1}(\mathbf{x}^{i}-\mathbf{x}^{j})\big)+\delta_{ij}\sigma_{\eta}^{2}$
in the following. Here, $\sigma_{f}^{2}$ and $\sigma_{\eta}^{2}$
denote the signal and noise variance, respectively and the diagonal
matrix $\mathbf{W}$ contains the squared length scales.
Since a GP is a distribution over functions $f:\mathcal{X}\rightarrow\mathcal{Z}$,
the output $\mathbf{z}=f(\mathbf{x})$ is random even though the input
$\mathbf{x}$ is deterministic. In GP regression, a GP prior is combined
with training data $\{\mathbf{x}^{i},z^{i}\}_{i\in\mathcal{I}=\{1..N\}}$
into a GP posterior conditioned on the training data with 
mean $\mu_{*}(\mathbf{x}_{*})=\bm{\alpha}^{\top}\mathbf{k}_{*}$
and covariance
$\bar{k}(\mathbf{x}_{*},\mathbf{x}'_{*})=k(\mathbf{x}_{*},\mathbf{x}'_{*})-\mathbf{k}_{*}^{\top}\mathbf{K}^{-1}\mathbf{k}'_{*}$
where $ $$\mathbf{k}_{*}=[k(\mathbf{x}^{1},\mathbf{x}_{*}),..,k(\mathbf{x}^{N},\mathbf{x}_{*})]^{\top}$,
$\mathbf{K}=[k(\mathbf{x}^{i},\mathbf{x}^{j})]_{ij=1..N}$, $\bm{\Sigma}_{*}=[\bar{k}(\mathbf{x}^{i},\mathbf{x}^{j})]_{ij=1..N}$
and $\bm{\alpha}^{\top}=[z^{1},..,z^{N}]\mathbf{K}^{-1}$. Deterministic
inputs $\mathbf{x}$ lead to Gaussian outputs and Gaussian inputs
lead to non-Gaussian outputs whose first two moments can be computed
analytically \cite{Candela03propUncertaintyTR} for ARD covariance. Multivariate 
deterministic inputs $\mathbf{x}$
lead to spherical Gaussian outputs $\mathbf{z}$ and Gaussian inputs
$\mathbf{x}$ lead to non-Gaussian outputs $\mathbf{z}$ whose moments 
$(\bm{\mu}_{*},\bm{\Sigma}_{*})$ are given by:

\begin{center}
\resizebox{\columnwidth}{!}{\begin{tabular}{|l|l||l||l|l||l|}
\hline 
\multicolumn{3}{|l||}{$\mathbf{x}\sim\delta(\mathbf{x}_{*})\stackrel{f\sim\mathcal{GP}}{\longrightarrow}\mathbf{z}\sim\mathcal{N}(\bm{\mu}_{*},\sigma_{*}^{2}\mathbf{I})$} & \multicolumn{3}{l|}{$\mathbf{x}\sim\mathcal{N}(\mathbf{x}_{*},\mathbf{V}_{\mathbf{x}})\stackrel{f\sim\mathcal{GP}}{\longrightarrow}\mathbf{z}\stackrel{(\approx)}{\sim}\mathcal{N}(\bm{\mu}_{*},\bm{\Sigma}_{*})$}\tabularnewline[\doublerulesep]
\hline
$\bm{\mu}_{*}=\mathbf{A}^{\top}\tilde{\mathbf{k}}_{*}$ & \multicolumn{2}{l||}{$\sigma_{*}^{2}=k_{**}-\mathbf{k}_{*}^{\top}\mathbf{K}^{-1}\mathbf{k}_{*}\in[\sigma_{\eta}^{2},\sigma_{\eta}^{2}+\sigma_{f}^{2}]$} & $\bm{\mu}_{*}=\mathbf{A}^{\top}\tilde{\mathbf{k}}_{*}$ & \multicolumn{2}{l|}{$\bm{\Sigma}_{*}=\left(k_{**}-\textrm{tr}(\mathbf{K}^{-1}\hat{\mathbf{K}}_{*})\right)\mathbf{I}+\mathbf{A}^{\top}(\hat{\mathbf{K}}_{*}-\tilde{\mathbf{k}}_{*}\tilde{\mathbf{k}}_{*}^{\top})\mathbf{A}$}\tabularnewline[\doublerulesep]
\hline
\end{tabular}}
\par\end{center}

Here, $\mathbf{A}^{\top}=[\bm{\alpha}_{1},..,\bm{\alpha}_{D}]^{\top}=[\mathbf{z}^{1},..,\mathbf{z}^{N}]\mathbf{K}^{-1}$
and the quantities $\tilde{\mathbf{k}}_{*}=\mathbb{E}[\mathbf{k}]$
and $\hat{\mathbf{K}}_{*}=\mathbb{E}[\mathbf{k}\mathbf{k}^{\top}]$
denote expectations of $\mathbf{k}=\mathbf{k}(\mathbf{x})=[k(\mathbf{x}^{1},\mathbf{x}),..,k(\mathbf{x}^{N},\mathbf{x})]^{\top}$
w.r.t. the Gaussian input distribution $\mathcal{N}(\mathbf{x}|\mathbf{x}_{*},\mathbf{V}_{\mathbf{x}})$
that can readily be evaluated in closed form \cite{Candela03propUncertaintyTR}
as detailed in the Appendix. In the limit of $\mathbf{V}_{\mathbf{x}}\rightarrow\mathbf{0}$
we recover the deterministic case as $\tilde{\mathbf{k}}_{*}\rightarrow\mathbf{k}_{*}$,
$\hat{\mathbf{K}}_{*}\rightarrow\mathbf{k}_{*}\mathbf{k}_{*}^{\top}$ and
$\bm{\Sigma}_{*}\rightarrow\sigma_{*}^{2}\mathbf{I}$. Non-zero input 
variance $\mathbf{V}_{\mathbf{x}}$ results in full
non-spherical output covariance $\bm{\Sigma}_{*}$, even for
independent GPs because all the GPs are driven by the same
(uncertain) input.

A GPLVM \cite{Lawrence05GPLVM}
is a successful and popular non-parametric Bayesian tool for high
dimensional nonlinear data modeling taking into account
the data's manifold structure based on a low-dimensional representation.
High dimensional data points $\mathbf{z}^{i}\in\mathcal{Z}\subset\mathbb{R}^{D}$,
$\mathbf{Z}=[\mathbf{z}^{1},\dots,\mathbf{z}^{N}]$, are represented
by corresponding latent points $\mathbf{X}=[\mathbf{x}^{1},\dots,\mathbf{x}^{N}]$
from a low-dimensional latent space $\mathcal{X}\subset\mathbb{R}^{d}$
mapped into $\mathcal{Z}$ by $D$ independent GPs $f_{j}$ -- one
for each component $z_{j}$ of the data. All the GPs $f_{j}$ are
conditioned on $\mathbf{X}$ and share the same covariance and mean
functions. The model is trained by maximising the sum of the log marginal
likelihoods over the $D$ independent regression problems with respect
to the latent points $\mathbf{X}$.

\begin{figure}[t]
\begin{centering}
\includegraphics[width=1\textwidth]{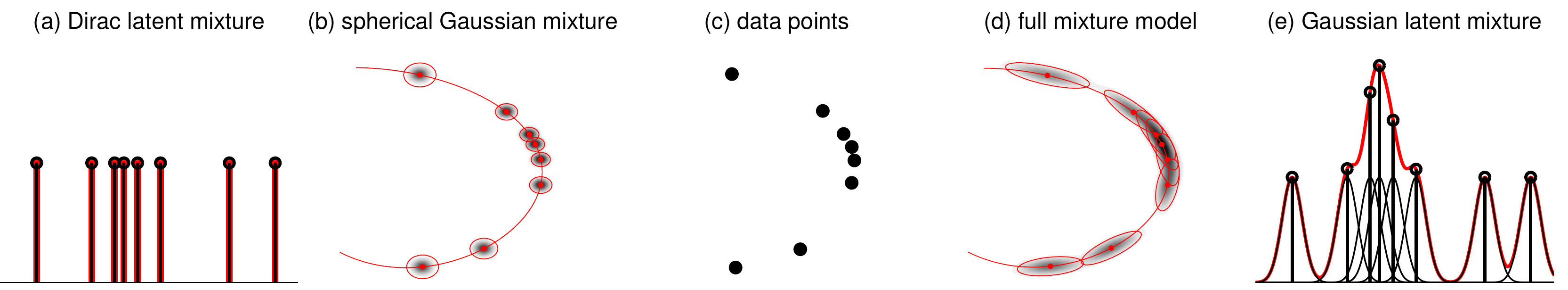}
\par\end{centering}

\caption{\label{fig:DGPLVM}The high dimensional ($D=2$) density of the data
points $\mathbf{z}^{i}$ in panel (c) is modelled by a mixture of
Gaussians $\mathbb{P}(\mathbf{z})=\frac{1}{N}\sum_{i}\mathcal{N}(\mathbf{z}|\bm{\mu}_{\mathbf{x}^{i}},\bm{\Sigma}_{\mathbf{x}^{i}})$
shown in panel (b,d) where the means $\bm{\mu}_{\mathbf{x}^{i}}$
and variances $\bm{\Sigma}_{\mathbf{x}^{i}}$ are given by the predictive
means and covariances of a set of independent Gaussian processes $f_{j}:\mathcal{X}\rightarrow\mathcal{Z}$
conditioned on low-dimensional ($d=1$) latent locations $\mathbf{x}^{i}$.
A latent Dirac mixture (a) yields a spherical Gaussian mixture with
varying widths (b) and a latent Gaussian mixture (e) results in a
fully coupled mixture model (d) smoothly sharing covariances across 
mixture components.}

\end{figure}

While most often applied to nonlinear dimensionality reduction, the
GPLVM can also be used as a tractable and flexible density model in
high dimensional spaces as illustrated in Figure \ref{fig:DGPLVM}.
The basic idea is to interpret the latent points $\mathbf{X}$ as
centres of a mixture of either Dirac (Figure \ref{fig:DGPLVM}a) or
Gaussian (Figure \ref{fig:DGPLVM}e) distributions in the latent space
$\mathcal{X}$ that are \emph{projected forward} by the GP  to produce
a high dimensional Gaussian mixture $\mathbb{P}(\mathbf{z})=\frac{1}{N}\sum_{i}\mathcal{N}(\mathbf{z}|\bm{\mu}_{\mathbf{x}^{i}},\bm{\Sigma}_{\mathbf{x}^{i}})$
in the observed space $\mathcal{Z}$. Depending on the kind of latent
mixture, the density model $\mathbb{P}(\mathbf{z})$ will either be
a mixture of spherical (Figure \ref{fig:DGPLVM}b) or full-covariance
Gaussians (Figure \ref{fig:DGPLVM}d). By that mechanism, we get a
tractable high dimensional density model $\mathbb{P}(\mathbf{z})$:
A set of low-dimensional coordinates in conjunction with a probabilistic
map $f$ yield a mixture of high dimensional Gaussians whose covariance
matrices are smoothly shared between components. As shown in Figure
\ref{fig:DGPLVM}(d), the model is able to capture high dimensional
covariance structure along the data manifold by relatively few parameters
(compared to $D^{2}$), namely the latent coordinates $\mathbf{X}\in\mathbb{R}^{d\times N}$
and the hyperparameters 
$\bm{\theta}=[\mathbf{W},\sigma_{f},\sigma_{\eta},\mathbf{V}_{\mathbf{x}}]\in\mathbb{R}_{+}^{2d+2}$
of the GP.

The role of the latent coordinates $\mathbf{X}$
is twofold: they both \emph{define the GP}, mapping the latent points
into the observed space, and they \emph{serve as centres} of the mixture
density in the latent space. If the latent density is a mixture of
Gaussians, the centres of these Gaussians are used to define the GP
map, but the full Gaussians (with covariance $\mathbf{V}_{\mathbf{x}}$)
are projected forward by the GP map.

\section{Learning Algorithm\label{sec:learning}}

Learning or model fitting is done by minimising a loss function $L$
w.r.t. the latent coordinates $\mathbf{X}$ and the hyperparameters
$\bm{\theta}$. In the following, we will discuss the usual GPLVM
objective function, make clear that it is not suited for density
estimation and use leave-out estimation to avoid overfitting.

\subsection{GPLVM likelihood \label{sec:learning_Z}}

A GPLVM \cite{Lawrence05GPLVM} is trained by setting the
latent coordinates $\mathbf{X}$ and the hyperparameters $\bm{\theta}$
to maximise the probability of the data
\vspace{-3mm}\begin{equation}
\mathbb{P}(\mathbf{Z}|\mathbf{X},\bm{\theta}) = \prod_{j=1}^{D}\mathbb{P}(\mathbf{z}_{j}|\mathbf{X},\bm{\theta}) = -\frac{DN}{2}\ln2\pi - \frac{D}{2}\ln\left|\mathbf{K}\right| - \frac{1}{2}\textrm{tr}\left(\mathbf{K}^{-1}\mathbf{Z}^{\top}\mathbf{Z}\right)\label{eq:GPLVM-Lik}
\end{equation} \vspace{-3mm} \\
that is the product of the marginal likelihoods of $D$ independent 
regression problems. Using 
$\frac{\partial L}{\partial\mathbf{K}}\negthickspace=\negthickspace\frac12\mathbf{K}^{-1}(\mathbf{Z}^{\top}\mathbf{Z}\negmedspace-\negmedspace D\mathbf{K})\mathbf{K}^{-1}$,
conjugate gradients optimisation at a cost of $\mathcal{O}(DN^{3})$ per step is
straightforward but suffers from local optima.

However, optimisation of $L_{Z}(\mathbf{X},\bm{\theta})$ does not
encourage the GPLVM to be a good density model. Only indirectly, we
expect the predictive variance to be small (implying high density)
in regions supported by many data points. The main focus of $L_{Z}(\mathbf{X},\bm{\theta})$
is on faithfully predicting $\mathbf{Z}$ from $\mathbf{X}$ (as implemented
by the fidelity trace term) while using a relatively smooth function
(as favoured by the log determinant term). Therefore, we propose a
different cost function.

\subsection{General leave-out estimators\label{sec:learning_generalLO}}

Density estimation \cite{wasserman06nonparamStat} constructs parametrised 
estimators $\hat{\mathbb{P}}_{\bm{\theta}}(\mathbf{z})$ from iid data 
$\mathbf{z}^{i}\sim\mathbb{P}(\mathbf{z})$. We use the Kullback-Leibler divergence 
$J(\bm{\theta})\stackrel{\text{c}}{=}-\int\mathbb{P}(\mathbf{z})\ln\hat{\mathbb{P}}_{\bm{\theta}}(\mathbf{z})\text{d}\mathbf{z}$
to the underlying density and its empirical estimate
$\hat{J}_{e}(\bm{\theta}) = -\sum_{i\in I}\ln\hat{\mathbb{P}}_{\bm{\theta},I}(\mathbf{z}^{i})$
as quality measure where $I$ emphasises that the full dataset has been
used for training. This estimator, is prone to overfitting if used to adjust 
the parameters via $\bm{\theta}^{*}=\arg\min_{\bm{\theta}}\hat{J}_{e}(\bm{\theta})$.
Therefore, estimators based on $K$ subsets of the data
$\hat{J}_{v}(\bm{\theta}) = -\frac{1}{K}\sum_{k=1}^{K}\sum_{i\notin I_{k}}\ln\hat{\mathbb{P}}_{\bm{\theta},I_{k}}(\mathbf{z}^{i}),\quad I_{k}\subset I$
are used. Two well known instances are $K$-fold cross-validation
(CV) and leave-one-out (LOO) estimation. The subsets for CV are
$I_{k}\cap I_{k'}=\emptyset,I=\bigcup_{k=1}^{K}I_{k},|I_{k}|\approx|I_{k'}|$
and $K=N,I_{k}=I\backslash\{k\}$ for LOO. Both of them can
be used to optimise $\bm{\theta}$.

\subsection{GPLVM leave-one-out density \label{sec:learning_LOO}}

There are two reasons why training a GPLVM with the log likelihood
of the data $L_{Z}(\mathbf{X},\bm{\theta})$ (Eq. \ref{eq:GPLVM-Lik})
is not optimal in the setting of density estimation: Firstly, it treats
the task as regression, and doesn't explicitly worry about how the
density is spread in the observation space. Secondly, our empirical
results (see Section \ref{sec:experiments}) indicate, that the test
set performance is simply not good.
Therefore, we propose to train the model using the leave-out density
\vspace{-3mm}\begin{equation}
-L_{LOO}(\mathbf{X},\bm{\theta})=\ln\prod_{i=1}^{N}\mathbb{P}_{\neg i}(\mathbf{z}^{i})=\sum_{i=1}^{N}\ln\frac{1}{N-1}\sum_{j\neq i}\mathcal{N}\left(\mathbf{z}^{i}|\bm{\mu}_{\mathbf{x}^{j}},\bm{\Sigma}_{\mathbf{x}^{j}}\right).\label{eq:DGPLVM-LOO}
\end{equation} \vspace{-3mm} \\ 
This objective is very different from the GPLVM criterion
as it measures how well a data point is explained under the mixture
models resulting from projecting each of the latent mixture components
forward; the leave-out aspect enforces that the point $\mathbf{z}^{i}$
gets assigned a high density even though the mixture component $\mathcal{N}\left(\mathbf{z}^{i}|\bm{\mu}_{\mathbf{x}^{i}},\bm{\Sigma}_{\mathbf{x}^{i}}\right)$
has been removed from the mixture. The leave-one-out idea is trivial
to apply in a mixture setting by just removing the contribution in
the sum over components, and is motivated by the desire to avoid overfitting.
Evaluation of $L_{LOO}(\mathbf{X},\bm{\theta})$ requires 
$\mathcal{O}(DN^{3})$ assuming $N>D>d$. 

However, removing the
mixture component is not enough since the latent point $\mathbf{x}^{i}$
is still present in the GP. Using rank one updates to compute inverses
and determinants of covariance matrices $\mathbf{K}_{\neg i}$
with row and column $i$ removed, it is possible to evaluate Eq. \ref{eq:DGPLVM-LPO}
for mixture components $\mathcal{N}\left(\mathbf{z}^{i}|\bm{\mu}_{\mathbf{x}^{j}}^{\neg i},\bm{\Sigma}_{\mathbf{x}^{j}}\right)$
with latent point $\mathbf{x}^{i}$ removed from the
mean prediction $\bm{\mu}_{\mathbf{x}^{j}}^{\neg i}$ -- which is
what we do in the experiments. Unfortunately, going further by removing
$\mathbf{x}^{i}$ also from the covariance $\bm{\Sigma}_{\mathbf{x}^{j}}$
increases the computational burden to $\mathcal{O}(DN^{4})$ because
we need to compute rank one corrections to all matrices $\hat{\mathbf{K}}_{\ell},\ \ell=1..N$.
Since $\bm{\Sigma}_{\mathbf{x}^{j}}^{\neg i}$
is only slightly smaller than $\bm{\Sigma}_{\mathbf{x}^{j}}$,
we refrain from computing it in the experiments.

In the original GPLVM, there is a clear one-to-one relationship between
latent points $\mathbf{x}^{i}$ and data points $\mathbf{z}^{i}$
-- they are inextricably tied together. However, the leave-one-out
(LOO) density $L_{LOO}(\mathbf{X},\bm{\theta})$ does not impose any
constraint of that sort. The number of mixture components does not
need to be $N$, in fact we can choose any number we like. Only the
data visible to the GP $\{\mathbf{x}^{j},\bar{\mathbf{z}}^{j}\}$
is tied together. The actual latent mixture centres $\mathbf{X}$
are not necessarily in correspondence with any actual data point $\mathbf{z}^{i}$.
However, we can choose $\bar{\mathbf{Z}}$ to be a subset of $\mathbf{Z}$.
This is reasonable because any mixture centre $\bm{\mu}_{\mathbf{x}^{j}}=\bar{\mathbf{Z}}\mathbf{K}^{-1}\mathbf{k}(\mathbf{x}^{j})$
(corresponding to the latent centre $\mathbf{x}^{j}$) lies in the
span of $\bar{\mathbf{Z}}$, hence $\bar{\mathbf{Z}}$ should approximately
span $\mathbf{Z}$. In our experiments, we enforce $\bar{\mathbf{Z}}=\mathbf{Z}$.

\subsection{Overfitting avoidance \label{sub:Overfit}}

Overfitting in density estimation means that very high densities are
assigned to training points, whereas very low densities remain for
the test points. Despite its success in parametric models, the leave-one-out
idea alone, is not sufficient to prevent overfitting in our model.
When optimising $L_{LOO}(\mathbf{X},\bm{\theta})$ w.r.t. $(\mathbf{X},\bm{\theta})$
using conjugate gradients, we observe the following
behaviour: The model circumvents the LOO objective by arranging the
latent centres in pairs that take care of each other. More generally,
the model partitions the data $\mathbf{Z}\subset\mathbb{R}^{D}$ into
groups of points lying in a subspace of dimension $\le D-1$ and adjusts
$(\mathbf{X},\bm{\theta})$ such that it produces a Gaussian with
very small variance $\sigma_{\perp}^{2}$ in the orthogonal complement
of that subspace. By scaling $\sigma_{\perp}^{2}$ to tiny values,
$L_{LOO}(\mathbf{X},\bm{\theta})$ can be made almost arbitrarily
large. It is understood that the hyperparameters of the underlying
GP take very extreme values: the noise variance $\sigma_{\eta}^{2}$
and some length scales $w_{i}$ become tiny. In $L_{Z}(\mathbf{X},\bm{\theta})$,
this is penalised by the $\ln|\mathbf{K}|$ term, but $L_{LOO}(\mathbf{X},\bm{\theta})$
is happy with very improbable GPs. In our initial experiments, we
observed this {}``cheating behaviour'' on several of datasets.

We conclude that even though the LOO objective (Eq. \ref{eq:DGPLVM-LPO})
is the standard tool to set KDE kernel widths \cite{wasserman06nonparamStat},
it breaks down for too complex models. We counterbalance this behaviour
by leaving out not only one point but rather $P$ points at a time.
This renders cheating tremendously difficult. In our experiments we
use the leave-$P$-out (LPO) objective
\vspace{-3mm}\begin{equation}
L_{LPO}(\mathbf{X},\bm{\theta})=-\sum_{k=1}^{K}\sum_{i\notin I_{k}}\ln\frac{1}{N-P}\sum_{j\in I_{k}}\mathcal{N}\left(\mathbf{z}^{i}|\bm{\mu}_{\mathbf{x}^{j}}^{\neg i},\bm{\Sigma}_{\mathbf{x}^{j}}\right).\label{eq:DGPLVM-LPO}
\end{equation} \vspace{-2mm} \\ 
Ideally, one would sum over all $K={N \choose P}$ subsets $I_{k}\in I$
of size $|I_{k}|=N-P$. However, the number of terms $K$ soon becomes
huge: $K\approx N^{P}$ for $P\ll N$. Therefore, we use an approximation
where we set $K=N$ and $I_{k}$ contains the indices $j$ that currently
have the smallest value $\mathcal{N}\left(\mathbf{z}^{k}|\bm{\mu}_{\mathbf{x}^{j}}^{\neg i},\bm{\Sigma}_{\mathbf{x}^{j}}\right)$.

All gradients $\frac{\partial L_{LPO}}{\partial\mathbf{X}}$ and $\frac{\partial L_{LPO}}{\partial\bm{\theta}}$
can be computed in $\mathcal{O}(DN^{3})$ when using $\bm{\mu}_{\mathbf{x}^{j}}^{\neg i}$.
However, the expressions take several pages. We use
a conjugate gradient optimiser to find the best parameters $\mathbf{X}$
and $\bm{\theta}$.

\section{Experiments\label{sec:experiments}}

In the experimental section, we show that the GPLVM trained with $L_{Z}(\mathbf{X},\bm{\theta})$
(Eq. \ref{eq:GPLVM-Lik}) does not lead to a good density model in
general. Using our $L_{LPO}$ training procedure (Section \ref{sub:Overfit},
Eq. \ref{eq:DGPLVM-LPO}), we can turn it into a competitive density
model. We demonstrate that a latent variance $\mathbf{V}_{\mathbf{x}}\succ\mathbf{0}$
improves the results even further in some cases and that on some datasets,
our density model training procedure performs better than all the
baselines.

\subsection{Datasets and baselines}

We consider $9$ data sets\footnote{\url{http://www.csie.ntu.edu.tw/~cjlin/libsvmtools/datasets/}}, 
frequently used in machine learning.
The data sets differ in their domain of application, their dimension $D$,
their number of instances $N$ and come from regression and classification. 
In our experiments, we do not use the labels.

\begin{center}
\resizebox{\columnwidth}{!}{\begin{tabular}{|c|c|c|c|c|c|c|c|c|c|}
\hline 
dataset & \texttt{breast} & \texttt{crabs} & \texttt{diabetes} & \texttt{ionosphere} & \texttt{sonar} & \texttt{usps} & \texttt{abalone} & \texttt{bodyfat} & \texttt{housing}\tabularnewline
\hline 
$N$,$D$ & \texttt{$449$,$9$} & \texttt{$200$,$6$} & \texttt{$768$,$8$} & \texttt{$351$,}$33$ & \texttt{$208$,$60$} & \texttt{$9298$,$256$} & \texttt{$4177$,$8$} & \texttt{$252$,$14$} & \texttt{$506$,$13$}\tabularnewline
\hline
\end{tabular}}
\par\end{center}

We do not only want to demonstrate that our training procedure yields
better test densities for the GPLVM. We are rather interested in a
fair assessment of how competitive the GPLVM is in density estimation
compared to other techniques. As baseline methods, we concentrate
on three standard algorithms: penalised fitting of a mixture of full
Gaussians (\texttt{gm}), kernel density estimation (\texttt{kde})
and manifold Parzen windows \cite{vincent03manifold} \texttt{(mp)}.
We run these algorithms for three different type of preprocessing:
raw data (r), data scaled to unit variance (\texttt{s}) and whitened
data (\texttt{w}). We explored a large number of parameter settings
and report the best results in Table \ref{tab:baselines}.

\subsubsection{Penalised Gaussian mixtures}

In order to speed up EM computations, we partition the dataset into
$K$ disjoint subsets using the $K$-means algorithm\texttt{}%
\footnote{\url{http://cseweb.ucsd.edu/~elkan/fastkmeans.html}%
}. We fitted a penalised Gaussian to each subset and combined them
using the relative cluster size as weight $\mathbb{P}(\mathbf{z})=\frac{1}{N}\sum_{k}N_{k}\mathbb{P}_{k}(\mathbf{z})$.
Every single Gaussian $\mathbb{P}_{k}(\mathbf{z})$ has the form $\mathbb{P}_{k}(\mathbf{z})=\mathcal{N}(\mathbf{z}|\mathbf{m}^{k},\mathbf{C}^{k}+w\mathbf{I})$
where $\mathbf{m}^{k}$ and $\mathbf{C}^{k}$ equal the sample mean
and covariance of the particular cluster, respectively. The global
ridge parameter $w$ prevents singular covariances and is chosen to
maximise the LOO log density
$ -L(w)=\ln\prod_{j}\mathbb{P}_{\neg j}(\mathbf{z}^{j})=\ln\prod_{j}\sum_{k}N_{k}\mathcal{N}(\mathbf{z}^{j}|\mathbf{m}_{\neg j}^{k},\mathbf{C}_{\neg j}^{k}+w\mathbf{I})$.
We use simple gradient descent to find the best parameter $w\in\mathbb{R}_{+}$.

\subsubsection{Diagonal Gaussian KDE}

The kernel density estimation procedure fits a mixture model
by centring one mixture component at each data point $\mathbf{z}^{i}$. 
We use independent multi-variate Gaussians:
$\mathbb{P}(\mathbf{z})=\frac{1}{N}\sum_{i}\mathcal{N}(\mathbf{z}|\mathbf{z}^{i},\mathbf{W})$,
where the diagonal widths $\mathbf{W}=\text{Dg}(w_{1},..,w_{D})$ are chosen to maximise the LOO density
$-L(\mathbf{W}) = \ln\prod_{j}\mathbb{P}_{\neg j}(\mathbf{z}^{j})=\ln\prod_{j}\frac{1}{N}\sum_{i\neq j}\mathcal{N}(\mathbf{z}^{j}|\mathbf{z}^{i},\mathbf{W})$.
We employ a Newton-scheme to find the best parameters $\mathbf{W}\in\mathbb{R}_{+}^{D}$.

\subsubsection{Manifold Parzen windows}

The manifold Parzen window estimator \cite{vincent03manifold} tries
to capture locality by means of a kernel $k$. It is a mixture
of $N$ full Gaussians where the covariance $\bm{\Sigma}^{i}=w\mathbf{I}+( \sum_{j\neq i}k(\mathbf{z}^{i},\mathbf{z}^{j})(\mathbf{z}^{i}-\mathbf{z}^{j})(\mathbf{z}^{i}-\mathbf{z}^{j})^{\top} )/(\sum_{j\neq i}k(\mathbf{z}^{i},\mathbf{z}^{j}))$
of each mixture component is only computed based on neighbouring
data points.

As proposed by the authors, we use the $r$-nearest neighbour kernel
and do not store full covariance matrices $\bm{\Sigma}^{i}$ but a
low rank approximation $\bm{\Sigma}^{i}\approx w\mathbf{I}+\mathbf{V}\mathbf{V}^{\top}$
with $\mathbf{V}\in\mathbb{R}^{D\times d}$. As in the other baselines, the ridge
parameter $w$ is set to maximise the LOO density.

\subsubsection{Baseline results}

\begin{table}[t]
\begin{centering}
\resizebox{\textwidth}{!}{
\begin{tabular}{|c|c|c|c|c|c|c|c|c|c|}
\hline
dataset & \texttt{breast} & \texttt{crabs} & \texttt{diabetes} & \texttt{ionosphere} & \texttt{sonar} & \texttt{usps} & \texttt{abalone} & \texttt{bodyfat} & \texttt{housing}\tabularnewline
\hline
\hline
$N_{tr}=50$ & $-9.1$ gm(10,s) & $0.9$ gm(5,r) & $-11.0$ gm(4,r) & $-34.1$ gm(10,r) & $-67.7$ gm(1,r) & $18.4$ gm(1,r) & $12.5$ gm(8,r) & $-36.0$ gm(1,w) & $-33.4$ gm(6,s)\tabularnewline
\hline
$N_{tr}=100$ & $-8.6$ gm(4,r) & $1.9$ gm(7,r) & $-10.0$ gm(3,w) & $-30.5$ gm(13,r) & $-62.0$ gm(1,r) & $124.8$ gm(1,r) & $13.9$ gm(5,r) & $-35.2$ gm(2,w) & $-30.6$ mp(6,21,s)\tabularnewline
\hline
$N_{tr}=150$ & $-8.4$ gm(9,s) &  & $-9.6$ gm(3,w) & $-33.9$ gm(6,s) & $-61.5$ gm(4,w) & $185.4$ gm(1,r) & $14.3$ gm(10,r) & $-34.7$ gm(5,w) & $-29.1$ mp(6,32,s)\tabularnewline
\hline
$N_{tr}=200$ & $-8.2$ gm(9,r) &  & $-8.3$ gm(4,w) & $-31.8$ gm(6,w) &  & $232.6$ gm(3,w) & $14.2$ gm(13,r) &  & $-23.5$ gm(4,s)\tabularnewline
\hline
$N_{tr}=250$ & $-8.1$ gm(11,r) &  & $-8.2$ gm(5,w) &  &  & $261.6$ gm(6,w) & $14.3$ gm(13,r) &  & $-16.0$ gm(3,w)\tabularnewline
\hline
\end{tabular}
}
\par\end{centering}

\caption{\label{tab:baselines} Average log test densities over $10$
random splits of the data. We did not allow $N_{tr}$ to
exceeded $N/2$. We only show the method yielding the highest test
density among the three baseline candidates penalised full Gaussian
mixture \texttt{gm}($K$,\texttt{$\rho$}), diagonal Gaussian kernel
density estimation \texttt{kde($\rho$)} and manifold Parzen windows
\texttt{mp}($d$,$r$,\texttt{$\rho$}). The parameter $K=\{1,..,13\}$
is the number of cluster centers used for \texttt{gm}, $d=\left\lceil D\cdot\{5,12,19,26,33,40\}/100\right\rceil $
is the number of latent dimensions and $r=\left\lceil N\cdot\{5,10,15,20,25,30\}/100\right\rceil $
the neighbourhood size for \texttt{mp} and $\rho$ is saying which
preprocessing has been used (raw \texttt{r}, scaled to unit variance
\texttt{s}, whitened \texttt{w}). The Gaussian mixture model yields
in all cases the highest test density except for one case where the
Parzen window estimator performs better.
}

\end{table}

The results of the baseline density estimators can be found in Table
\ref{tab:baselines}. They clearly show three things: (i) More data
yields better performance, (ii) penalised mixture of Gaussians is
clearly and consistently the best method and (iii) manifold Parzen
windows \cite{vincent03manifold} offer only little benefit. The
absolute values can only be compared within datasets since linearly
transforming the data $\mathbf{Z}$ by $\mathbf{P}$ results in a constant
offset $\ln|\mathbf{P}|$ in the log test probabilities.

\subsection{Experimental setting and results}

\begin{figure}[t]
\begin{centering}
\includegraphics[width=1\textwidth]{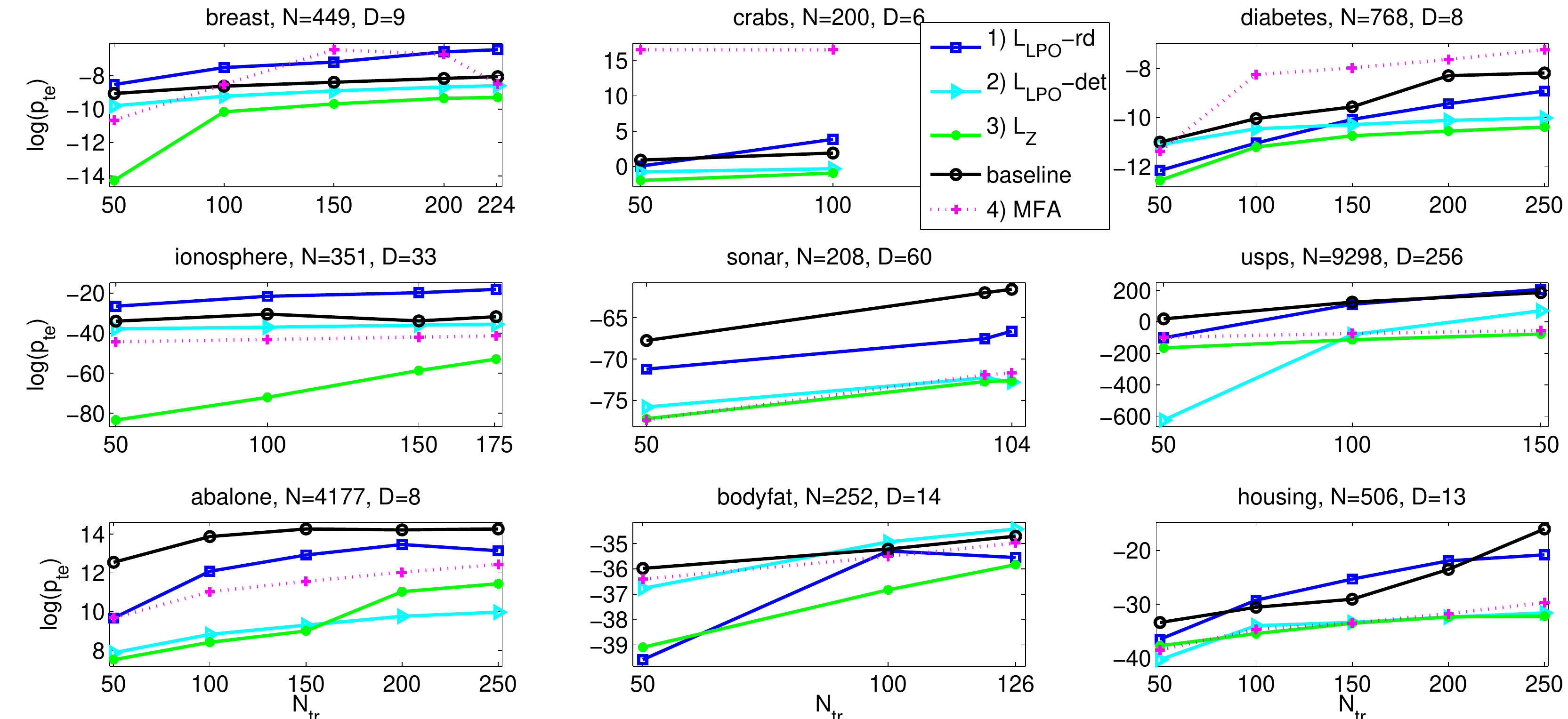}
\par\end{centering}

\caption{\label{fig:test_density}Each panel displays the log test density 
averaged over $10$ random splits for three different GPLVM training 
procedures and the \emph{best} out of 41 baselines (penalised mixture
$k=1..13$, diag.+isotropic KDE, manifold Parzen windows with 36 different
parameter settings) as well as various mixture of factor analysers (MFA)
settings as a function of the number of training data points
$N_{tr}$.
We report the maximum value across latent dimension $d=\{1,2,3\}$,
three preprocessing methods (raw, scaled to unit variance, whitened)
and $P=\{1,2,5,10,15\}$ leave-out points . The GPLVM training
procedures are the following: $L_{LPO}$-rd: stochastic leave-$P$-out
density (Eq. \ref{eq:DGPLVM-LPO} with latent Gaussians, $\mathbf{V}_{\mathbf{x}}\succ\mathbf{0}$),
$L_{LPO}$-det: deterministic leave-$P$-out density (Eq. \ref{eq:DGPLVM-LPO}
with latent Diracs, $\mathbf{V}_{\mathbf{x}}=\mathbf{0}$) and $L_{Z}$:
marginal likelihood (Eq. \ref{eq:GPLVM-Lik}).}
\end{figure}

We keep the experimental schedule and setting of the previous Section
in terms of the $9$ datasets, the $10$ fold averaging procedure
and the maximal training set size $N_{tr}=N/2$. We use the GPLVM
log likelihood of the data $L_{Z}(\mathbf{X},\bm{\theta})$, the LPO
log density with deterministic latent centres ($L_{LPO}\texttt{-det}(\mathbf{X},\bm{\theta}),\:\mathbf{V}_{\mathbf{x}}=\mathbf{0}$)
and the LPO log density using a Gaussian latent centres $L_{LPO}\texttt{-rd}(\mathbf{X},\bm{\theta})$
to optimise the latent centres $\mathbf{X}$ and the hyperparameters
$\bm{\theta}$. Our numerical results include $3$ different latent
dimensions $d$, $3$ preprocessing procedures and $5$ different
numbers of leave-out points $P$. Optimisation is done using $600$
conjugate gradient steps alternating between $\mathbf{X}$ and $\bm{\theta}$.
In order to compress the big amount of numbers, we report the method
with highest test density as shown in Figure \ref{fig:test_density},
only.

The most obvious conclusion, we can draw from the numerical experiments,
is the bad performance of $L_{Z}(\mathbf{X},\bm{\theta})$ as a training
procedure for GPLVM in the context of density modeling. This finding
is consistent over all datasets and numbers of training points. We
get another conclusive result in terms of how the latent variance
$\mathbf{V}_{\mathbf{x}}$ influences the final test densities%
\footnote{In principle, $\mathbf{V}_{\mathbf{x}}$ could be
fixed to $\mathbf{I}$ because its scale can be modelled by $\mathbf{X}$.}. Only in the \texttt{bodyfat} data set it is not beneficial to allow
for latent variance. It is clear that this is an intrinsic property
of the dataset itself, whether it prefers to be modelled by a spherical
Gaussian mixture or by a full Gaussian mixture.

An important issue, namely how well a fancy density model performs
compared to very simple models, has in the literature either been
ignored \cite{bishop98GTM,Roweis02globLinMod} or only done
in a very limited way \cite{vincent03manifold}. Experimentally,
we can conclude that on some datasets e.g. \texttt{diabetes, sonar,
abalone} our procedure cannot compete with a plain \texttt{gm} model.
However note, that the baseline numbers were obtained as the maximum
over a wide (41 in total) range of parameters and methods.

For example, in the \texttt{usps} case, our elaborate density estimation
procedure outperforms a single penalised Gaussian only for training
set sizes $N_{tr}>100$. However, the margin in terms of density is
quite big: On $N_{tr}=150$ prewhitened data points $L_{LPO}(\mathbf{X},\bm{\theta})$
with deterministic latents yields $70.47$ at $d=2$, whereas full
$L_{LPO}(\mathbf{X},\bm{\theta})$ reaches $207$ at $d=4$ which
is significantly above $185.4$ as obtained by the \texttt{gm} method
-- since we work on a logarithmic scale, this corresponds to factor
of $2.4\cdot10^{9}$ in terms of density.

\subsection{Running times}

While the baseline methods such as\texttt{ gm,} \texttt{kde} and \texttt{mp}
run in a couple of minutes for the \texttt{usps} dataset, training
a GPLVM with either $L_{LPO}(\mathbf{X},\bm{\theta}),\:\mathbf{V}_{\mathbf{x}}=\mathbf{0}$
or $ $$L_{Z}(\mathbf{X},\bm{\theta})$ takes considerably longer
since a lot of cubic covariance matrix operations need to be computed
during the joint optimisation of $(\mathbf{X},\bm{\theta})$. The
GPLVM computations scale cubically in the number of data
points $N_{tr}$ used by the GP forward map and quadratically in the
dimension of the observed space $D$. The major computational gap
is the transition from $\mathbf{V}_{\mathbf{x}}=\mathbf{0}$ to $\mathbf{V}_{\mathbf{x}}\succ\mathbf{0}$
because in the latter case, covariance matrices of size $D^{2}$ have
to be evaluated which cause the optimisation to last in the order
of a couple of hours. To provide concrete timing results, we picked
$N_{tr}=150$, $d=2$, averaged over the $9$ datasets and show times
relative to $L_{Z}$.

\begin{center}
\resizebox{.7\columnwidth}{!}{
\begin{tabular}{|c|c|c|c|c|c|c|c|c|}
\hline
alg & \texttt{gm(1)} & \texttt{gm(10)} & \texttt{kde} & \texttt{mp} & \texttt{mfa} &\texttt{$L_{Z}$} & \texttt{$L_{LPO}$-det} & \texttt{$L_{LPO}$-rd}\tabularnewline
\hline
$t_{rel}$& $0.27$ & $0.87$ & $0.93$ & $1.38$ & $0.30$ & $1.00$ & $35.39$ & $343.37$ \tabularnewline
\hline
\end{tabular}
}
\par\end{center}

Note that the methods $L_{LPO}$ are run in a conservative fail-proof
black box mode with $600$ gradient steps. We observe good densities 
after considerably less
gradient steps already. Another straightforward speedup can be obtained by
carefully pruning the number of inputs to the $L_{LPO}$ models.

\section{Conclusion and Discussion\label{sec:discussion}}

We have discussed how the basic GPLVM is not in itself a good 
density model, and results on several datasets have shown, that
it does not generalise well. We have discussed two
alternatives based on explicitly projecting forward a mixture model
from the latent space. Experiments show that such density models are
generally superior to the simple GPLVM.

Among the two alternative ways of defining the latent densities, the
simplest is a mixture of delta functions, which -- due to the stochasticity
of the GP map -- results in a smooth predictive distribution.
However, the resulting mixture of Gaussians, has only axis aligned
components. If instead the latent distribution is a mixture of Gaussians,
the dimensions of the observations become correlated. This allows
the learnt densities to faithfully follow
the underlying manifold.

Although the presented model has attractive properties, some problems
remain: The learning algorithm needs a good initialisation and the
computational demand of the method is considerable. However, we have
pointed out that in contrast to the GPLVM, the number of latent points
need not match the number of observations allowing for alternative
sparse methods.

We have detailed how to adapt ideas based on the GPLVM to density modeling in 
high dimensions and have shown that such models
are feasible to train.

\vspace{-4mm}
{\small
\renewcommand{\baselinestretch}{0.95}         
\bibliographystyle{splncs}
\bibliography{density_gplvm_arxiv}}

\vspace{-4mm}
\subsubsection*{Appendix \label{sec:Appendix}}

Both
$\hat{\mathbf{K}}_{*}=\mathbb{E}[\mathbf{k}\mathbf{k}^{\top}]=[\hat{k}_{*}(\mathbf{x}^{i},\mathbf{x}^{j})]_{ij}$ 
and 
$\tilde{\mathbf{k}}_{*}=\mathbb{E}[\mathbf{k}]=[\tilde{k}_{*}(\mathbf{x}^{j})]_{j}$
are the following expectations of $\mathbf{k}=[k(\mathbf{x},\mathbf{x}^{1}),..,k(\mathbf{x},\mathbf{x}^{N})]^{\top}$
w.r.t. $\mathcal{N}(\mathbf{x}|\mathbf{x}_{*},\mathbf{V}_{\mathbf{x}})$:
\begin{eqnarray*}
\tilde{k}_{*}(\mathbf{x}^{i}) & = & \sigma_{f}^{2}\left|\mathbf{V}_{\mathbf{x}}\mathbf{W}^{-1}+\mathbf{I}\right|^{-\frac{1}{2}}\rho\left(\mathbf{V}_{\mathbf{x}}+\mathbf{W},\mathbf{x}^{i}-\mathbf{x}_{*}\right),\:  \rho(\mathbf{D},\mathbf{y})=e^{-\frac{1}{2}\mathbf{y}^{\top}\mathbf{D}^{-1}\mathbf{y}}, \:\text{and} \\
\hat{k}_{*}(\mathbf{x}^{i},\mathbf{x}^{j}) & = & \frac{k(\mathbf{x}^{i},\mathbf{x}_{*})k(\mathbf{x}^{j},\mathbf{x}_{*})}{\sqrt{\left|2\mathbf{V}_{\mathbf{x}}\mathbf{W}^{-1}+\mathbf{I}\right|}}\rho\left(\frac{1}{2}\mathbf{W}\mathbf{V}_{\mathbf{x}}^{-1}\mathbf{W}+\mathbf{W},\frac{\mathbf{x}^{i}+\mathbf{x}^{j}}{2}-\mathbf{x}_{*}\right).\end{eqnarray*}

\end{document}